\newcommand*\chemeleon{\textbf{CheMeleon}}
\newcommand*\rf{\textbf{RF}}
\newcommand*\mordred{Mordred}
\newcommand*\chemprop{\textbf{Chemprop}}
\newcommand*\minimol{\textbf{minimol}}
\newcommand*\molformer{\textbf{MoLFormer}}
\newcommand*\fastprop{\textbf{fastprop}}

\newcommand*\molclr{\textbf{MolCLR}}
\newcommand*\graphqpt{\textbf{GraphQPT}}
\newcommand*\molgps{\textbf{MolGPS}}

\newcommand*\chemberta{\textbf{ChemBERTa}}

\newcommand{\bigO}{\mathcal{O}}

\documentclass[sn-mathphys-num]{config/sn-jnl}
\usepackage{graphicx}%
\usepackage{multirow}%
\usepackage{amsmath,amssymb,amsfonts}%
\usepackage{amsthm}%
\usepackage{mathrsfs}%
\usepackage[title]{appendix}%
\usepackage{xcolor}%
\usepackage{textcomp}%
\usepackage{manyfoot}%
\usepackage{booktabs}%
\usepackage{ragged2e}
\usepackage{siunitx}%
\usepackage{algorithm}%
\usepackage{algorithmicx}%
\usepackage{algpseudocode}%
\usepackage{listings}%
\theoremstyle{thmstyleone}%
\theoremstyle{thmstyletwo}%
\theoremstyle{thmstylethree}%
\usepackage{threeparttable}
\usepackage{caption}
\usepackage{xr}
\usepackage{float}
\usepackage[section]{placeins}

\pagecolor{white}
\usepackage{pdfpages}

\begin{document}


\title[Article Title]{Deep Learning Foundation Models from Classical Molecular Descriptors}
\author[1]{\fnm{Jackson W.} \sur{Burns}}
\equalcont{These authors contributed equally to this work.}
\author[1]{\fnm{Akshat Shirish} \sur{Zalte}}
\equalcont{These authors contributed equally to this work.}
\author[1]{\fnm{Charlles R. A.} \sur{Abreu}}
\author[2]{\fnm{Jochen} \sur{Sieg}}
\author[2]{\fnm{Christian} \sur{Feldmann}}
\author[2]{\fnm{Miriam} \sur{Mathea}}
\author*[1]{\fnm{William H.} \sur{Green}}\email{whgreen@mit.edu}
\affil[1]{\orgdiv{Department of Chemical Engineering}, \orgname{MIT}, \orgaddress{\city{Cambridge}, \state{Massachusetts}, \country{USA}}}
\affil[2]{\orgdiv{\orgname{BASF SE}, \orgaddress{\city{Ludwigshafen}, \country{Germany}}}}


\abstract{
\unboldmath
Fast and accurate data-driven prediction of molecular properties is pivotal to scientific advancements across myriad chemical domains.
Deep learning methods have recently garnered much attention, despite their inability to outperform classical machine learning methods when tested on practical, real-world benchmarks with limited training data.
This study seeks to bridge this gap with \chemeleon{}, a $\bigO(10M)$ parameter foundation model that enables directed message-passing neural networks to finally exceed the performance of classical methods.
Evaluated on 58 benchmark datasets from Polaris and MoleculeACE, \chemeleon{} achieves a win rate of 75\% on Polaris tasks, outperforming baselines like Random Forest (68\%), \fastprop{} (36\%), and \chemprop{} (32\%), and a 97\% win rate on MoleculeACE assays, surpassing Random Forest (50\%) and other foundation models.
Unlike conventional pre-training approaches that rely on noisy experimental data or biased quantum mechanical simulations, \chemeleon{} utilizes low-noise molecular descriptors to learn rich and highly transferable molecular representations, suggesting a new avenue for foundation model pre-training.
}
\maketitle


Fast and accurate prediction of molecular properties is a shared challenge pivotal to advancements across diverse scientific domains.
In recent years, artificial intelligence (AI) and deep learning (DL) have emerged as transformative tools for this task.
By providing rapid surrogates for resource-intensive experiments or complex quantum mechanics (QM) simulations, these data-driven approaches offer the potential to significantly accelerate the iterative process of discovering molecules with desirable properties.
The success of these methods relies heavily on their ability to map molecular structures to their functional properties.
Historically, this challenge has been approached through two distinct paradigms: classical methods based on fixed representations, and modern deep learning approaches based on learned representations.

The field of cheminformatics has long relied on effective classical methods that utilize molecular representations derived from expert chemical knowledge.
Molecular fingerprints, such as Morgan fingerprints \cite{Morgan1965}, encode structural substructures into bit vectors, while molecular descriptors, like those from \mordred{} \cite{mordred}, provide dense summaries of topological and physicochemical properties.
Classical models train a Random Forest or feed-forward neural network (FNN) on these expert-crafted features and are robust baselines, often outperforming modern deep learning architectures on practical benchmarks where data is scarce \cite{deeplose, fastprop}.

In contrast, graph neural networks (GNNs), such as the directed message passing neural network (D-MPNN) \cite{chemprop_theory}, aim to learn optimal representations directly from the molecular graph.
This approach offers a flexible and generalizable framework that has been successfully applied to predict a wide range of molecular properties \cite{stokes2020deep, greenman2022multi, swanson2024admet, fastsolv, rigr, alibrahim2025accurately, liao2025directed}.
While theoretically more powerful, these learned-representation models struggle in the low-data regime ($\lesssim \bigO(1{,}000)$ training samples) typical of drug discovery campaigns.
Without large datasets to guide them, they fail to simultaneously learn meaningful molecular features and the complex correlations with target properties, often leading to poorer performance when compared to classical methods.

To bridge this gap and unlock the potential of deep learning for chemistry, the field has increasingly turned to foundation models.
Mirroring the revolution in natural language processing (NLP) and computer vision, driven by models like GPT \cite{gpt3}, these large neural networks are pre-trained on vast datasets to learn generalizable representations that can be fine-tuned for various prediction tasks.
A comprehensive review of models by \citeauthor{choi2025perspective} \cite{choi2025perspective} highlights the evolution and impact of these approaches.

Early efforts adapted techniques from NLP to textual representations of molecules, such as SMILES or SELFIES \cite{selfies}.
Models like \chemberta{} \cite{chemberta} and \molformer{} \cite{molformer} treat chemical strings as language, employing masked language modeling to learn syntax and grammar from vast compound collections.
Recent models have shifted toward graph-based architectures, which offer a stronger inductive bias for molecular topology.
These approaches typically employ self-supervised learning strategies such as contrastive learning \cite{molclr}, context prediction \cite{hu_strategies}, or motif-level prediction tasks \cite{grover, kermt, mole, motif_learning} to capture structural similarity and chemically meaningful substructures.
While promising, these methods often rely on proxy tasks that may not directly align with the physicochemical properties relevant to downstream applications.

Alternatively, supervised pre-training has emerged as a powerful framework.
Therein, models are pre-trained to predict specific molecular properties.
Recent models, such as \minimol{} \cite{minimol}, \molgps{} \cite{molgps}, and \graphqpt{} \cite{graphqpt}, demonstrate that pre-training on labeled data can yield powerful foundation models that often outperform their self-supervised counterparts.
However, this approach faces a critical bottleneck: the quality and availability of ground truth labels.
Experimental labels are often sparse and noisy, suffering from significant inter-laboratory variability \cite{landrum_combining}, which introduces both stochastic and systematic error into the pre-training process.
Labels derived from QM simulations, while more abundant, can be computationally expensive to generate at scale and are often prone to systematic biases or limited to specific regions of chemical space.
Consequently, existing foundation models must trade off between the scale of unlabeled data and the quality of labeled data.

In this study, we explore a third, underutilized source of ground truth: easy-to-calculate classical molecular descriptors.
Unlike experimental measurements or QM simulations, descriptors are deterministic, low-noise, and computationally inexpensive to generate for any valid molecular structure.
We hypothesize that pre-training a model to predict these classical descriptors compels it to internalize the rich, expert-derived chemical knowledge they contain, ranging from simple atom counts to complex topological indices and additive physicochemical property estimators.
This approach effectively combines the strengths of classical cheminformatics with modern deep learning: the model learns to construct a representation that encodes the dense information of descriptors, but does so within a flexible, differentiable neural network architecture that allows for end-to-end fine-tuning on specific downstream tasks.

\begin{figure*}
    \centering
    \includegraphics[width=1\linewidth]{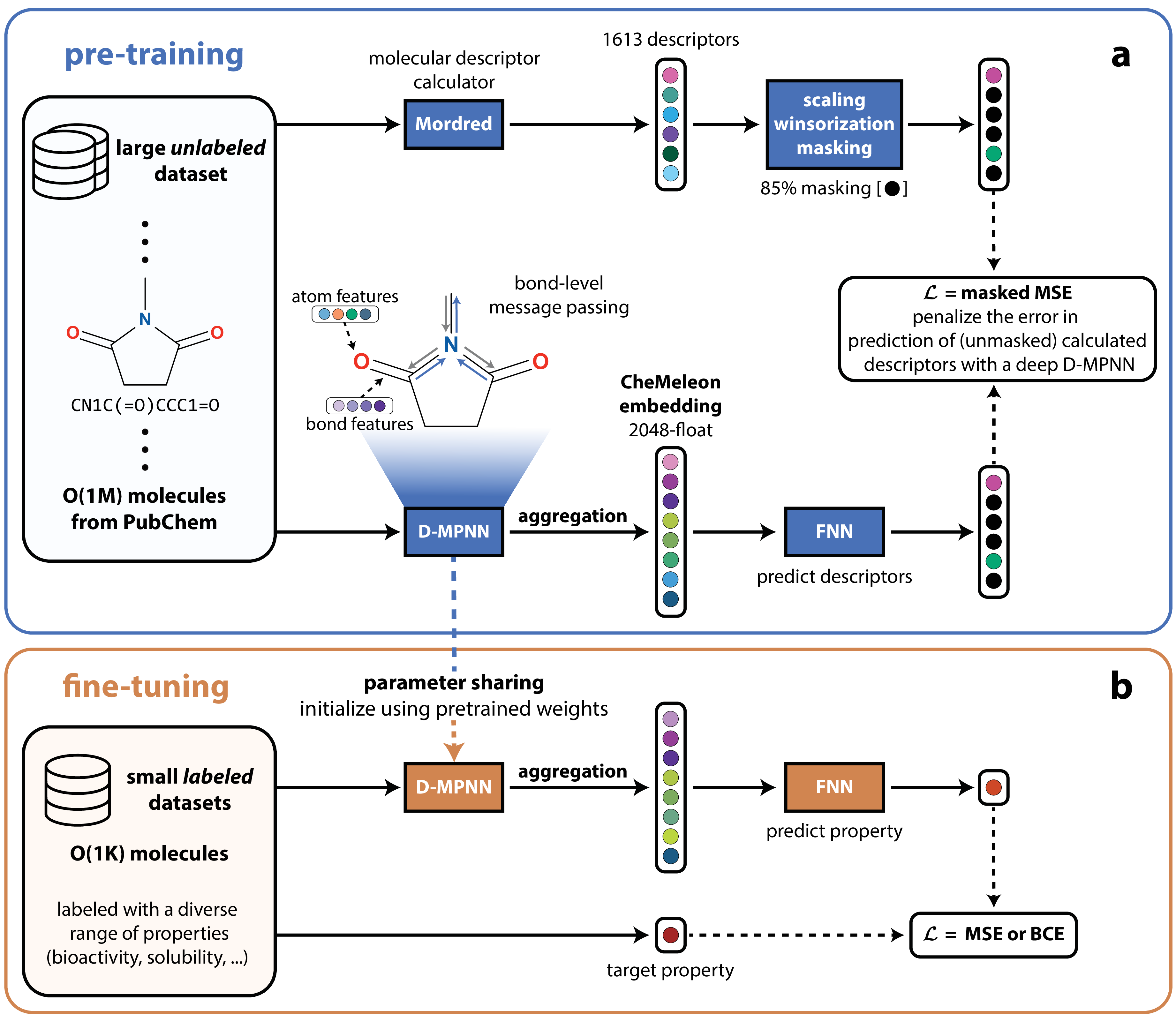}
    \caption{
    Workflow for the present study.
    (a) A large corpus of unlabeled SMILES are randomly selected from PubChem \cite{pubchem} and each is featurized into a vector of molecular descriptors using \mordred{} \cite{mordred}.
    \chemprop{} is used to train a directed message passing neural network \cite{chemprop_theory,chemprop_software} (D-MPNN) to predict these descriptors using a masked loss analogous to \chemberta{}'s as a form of regularization \cite{chemberta}.
    (b) The resulting D-MPNN is then reused for subsequent fine-tuning on smaller downstream datasets labeled with quantities of interest, such as bioactivity.
    }
    \label{fig:workflow}
\end{figure*}

\FloatBarrier

We introduce \chemeleon{}, a foundation model pre-trained on 1 million molecules from PubChem \cite{pubchem} to predict their computed \mordred{} descriptors using a D-MPNN architecture.
Figure \ref{fig:workflow} outlines the pre-training and fine-tuning workflow of \chemeleon{}.
By utilizing descriptors as targets, the model learns robust and chemically-informed representations free from the confounding effects of label noise.
With end-to-end fine-tuning, \chemeleon{} consistently outperforms strong classical baselines like Random Forest (\rf{}) as well as existing foundation models when benchmarked on 58 practically relevant tasks covering properties like solubility, lipophilicity, and bioactivity.
We also present an analysis of \chemeleon{}'s learned fingerprint, demonstrating its ability to capture chemically meaningful relationships and generalize across diverse molecular spaces.
Crucially, this state-of-the-art performance is achieved with an off-the-shelf model architecture and descriptor set, demonstrating the effectiveness of our pre-training method.

\section{Results}
\label{results}

We present summary statistics and key graphical results following the procedures described in Section~\ref{methods}. For clarity, only a representative subset is shown here, while the complete set of results and all analysis code are available in the Supporting Information and at the source code repository (see Section~\ref{code_avail}).

\subsection{Pre-training}
\label{results_pretraining}

\chemeleon{} was constructed using the D-MPNN architecture implemented in \chemprop{}.
Following \chemprop{} conventions, the message-passing portion of the network is of dimension 2048, containing 8.7 million parameters.
The number of message passing iterations (depth) was set to 6 and mean aggregation was used to yield the molecular embedding.
The subsequent two-layer FNN of the same dimension maps this representation to the molecular descriptors for a total parameter count of 12.9 million.
After pre-training, the test root mean squared error (RMSE) was 0.14, averaged across all 1613 \mordred{} descriptors, where each descriptor was rescaled to have zero mean and unit variance and Winsorized to six standard deviations.

\subsection{Polaris Benchmarks}
\label{results_polaris}

\begin{figure}[H]
    \centering
    \includegraphics[width=0.80\linewidth]{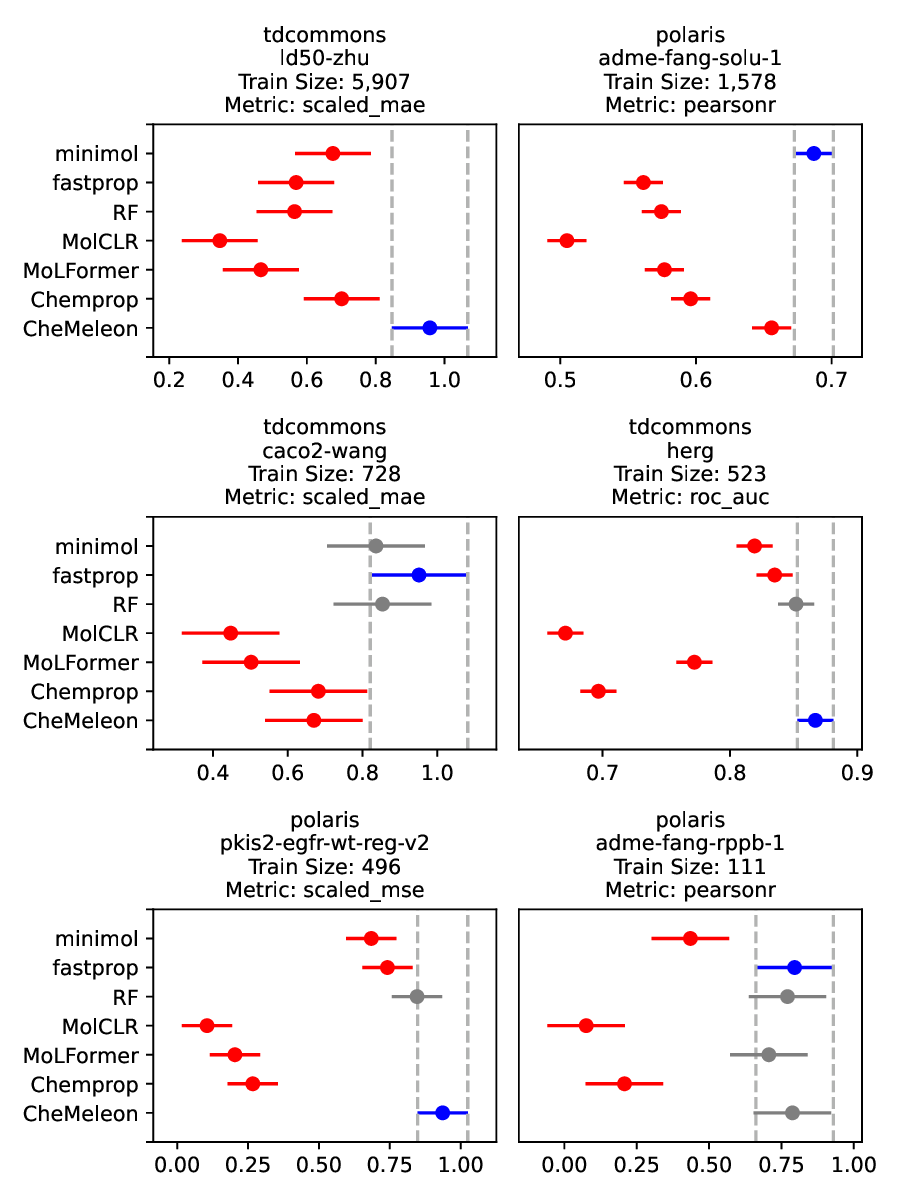}
    \caption{
    Performance of all of the tested models across a set of different common molecular machine learning tasks.
    The origin of each benchmark set is shown as the first line of each subplot title, followed by the name of the dataset (which indicates the task), the size of the training data, and the metric used to evaluate model performance.
    Benchmarks are sorted by training size in decreasing order.
    Models shown in blue are the absolute highest performers on the given benchmark, while models shown in gray are not practically different from the best performer according to the Tukey Honestly Significant Difference test ($\alpha=0.05$) based on the variance in test set performance across five repetitions, as laid out in Section \ref{benchmarks}.
    Models shown in red \textit{are} practically worse performers and are considered to have ``lost'' on the indicated benchmark.
    }
    \label{fig:polaris_hsd_selected}
\end{figure}

Figure \ref{fig:polaris_hsd_selected} shows the results for a select set of benchmarks covering particularly relevant tasks and a span of dataset sizes.
This diagram follows the conventions of \citeauthor{polaris} \cite{polaris}, with the best performing model according to the indicated metric shown in blue, statistically indistinguishable models shown in gray, and statistically inferior models shown in red (see Section \ref{methods} for further information).

For a complete diagram showing all of the Polaris benchmarks, see Supporting Information \ref{detailed_experimental_results}.
These results are summarized in Table \ref{tab:model_performance_polaris}.
Models achieving the best or statistically indistinguishable performance relative to the best are given a ``win'' on each benchmark, reflecting their overall performance across this range of relevant tasks.

\begin{table}[h]
\centering
\caption{Model performance comparison on Polaris benchmarks.}
\label{tab:model_performance_polaris}
\begin{tabular}{l c S[table-format=2.1]} 
\toprule
\textbf{Model} & \textbf{Win Count} & \textbf{Win Rate (\%)} \\
\midrule
\chemeleon{}        & 21        & 75        \\
\minimol{}             & 20        & 71          \\
\rf{} & 19 & 68 \\
\molformer{}           & 11        & 39          \\
\fastprop{}            & 10        & 36          \\
\chemprop{}            & 9        & 32         \\
\molclr{} & 4 & 14 \\
\bottomrule
\end{tabular}
\end{table}

\subsection{MoleculeACE Benchmarks}
\label{results_moleculeace}

\begin{figure}[H]
    \centering
    \includegraphics[width=1\linewidth]{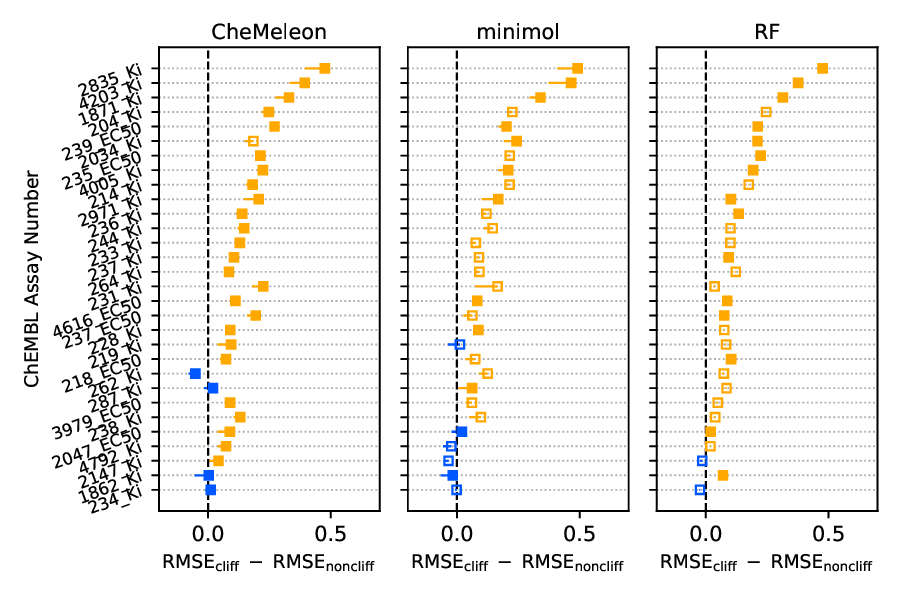}
    \caption{
    Performance of models across the ChEMBL assays \cite{chembl} curated as part of the MoleculeACE study \cite{moleculeace}.
    Each marker indicates the difference in Root Mean Squared Error (RMSE) of predictions between molecules in the cliff set and those not in the cliff set (noncliff), for the specified assay.
    Five-fold cross validation was performed according to the procedure described in Section \ref{statistical_comparisons}, enabling a one-sided t-test to check if this performance difference was statistically greater than zero (confidence interval for this test is shown as horizontal error bars, $\alpha=0.05$).
    Markers shown in blue are not practically different from zero, a positive result indicating that the model performance on the two sets is indistinguishable.
    Marker filling reflects the absolute performance of the models relative to one another following the same statistical procedure as Figure \ref{fig:polaris_hsd_selected}.
    Filled markers indicate that the given model was statistically the best or indistinguishable from the best performer in terms of RMSE for the entire test set; hollow markers indicate statistically significant worse performance.
    Absolute, overall performances results are also presented in Supporting Information Figure \ref{fig:mace_hsd_all} in the style of Figure \ref{fig:polaris_hsd_selected}.
    }
    \label{fig:mace_all}
\end{figure}

Figure \ref{fig:mace_all} shows the three best-performing models from the Polaris benchmark set evaluated on the MoleculeACE benchmark set.
This challenging task tests a model's capacity to consistently predict biological activity when small structural modifications cause unexpectedly large activity differences.

This figure is inspired by that in the original study \cite{moleculeace}, but modified to show both the per-model statistical tests or consistency and the intra-model test for absolute performance.
The RMSE difference between activity cliff subgroups is shown on the horizontal axis across all assays in the MoleculeACE study on the vertical axis.
Filled markers indicate the statistically best models (according to the procedure described in Figure \ref{fig:polaris_hsd_selected}) and blue and yellow indicate models which are consistent and inconsistent, respectively, between cliff and noncliff compounds.

Section \ref{methods} provides further information about the construction of these diagrams.
The results for all models are shown in Supporting Information Section \ref{detailed_experimental_results}.

Following the conventions of Table \ref{tab:model_performance_polaris}, summary statistics for all models performance on activity prediction are presented in Table \ref{tab:model_performance_moleculeace}.
Win Count and Rate are reported for the entire test set and the cliff set separately.

\begin{table}[h]
\centering
\caption{Model performance comparison on MoleculeACE benchmarks.}
\label{tab:model_performance_moleculeace}
\begin{tabular}{lcccc S[table-format=2.1]} 
\toprule
& \multicolumn{2}{c}{\textbf{Entire Test Set}} & \multicolumn{2}{c}{\textbf{Cliff Subset Only}} \\
\cmidrule(lr){2-3} \cmidrule(lr){4-5}
\textbf{Model} & \textbf{Win Count} & \textbf{Win Rate (\%)} & \textbf{Win Count} & \textbf{Win Rate (\%)} \\
\midrule
\chemeleon{}   & 29 & 97 & 30 & 100\\
\rf{} & 15 & 50 & 20 & 67 \\
\minimol{}        & 12 & 40 & 20 & 67 \\
\fastprop{}       &  5 & 17 & 12 & 40 \\
\molformer{}      &  5 & 17 & 10 & 33 \\
\chemprop{}       &  0 &  0 & 4 & 13 \\
\molclr{} & 0 & 0 & 2 & 7 \\
\bottomrule
\end{tabular}
\end{table}

\newpage

\subsection{k-Nearest Neighbors Representation Probing}
\label{knn_probing}

\chemeleon{}'s learned molecular representation was evaluated  through a set of k-Nearest Neighbors (kNN) probing experiments on 20 challenging toxicity classification endpoints \cite{richard2016toxcast}.
This evaluation is conceptually related to read-across approaches, in which the known toxicity profiles of chemically similar compounds are used to support or refute toxicity predictions for a query molecule.

\begin{figure}[h]
    \centering
    \includegraphics[width=1\linewidth]{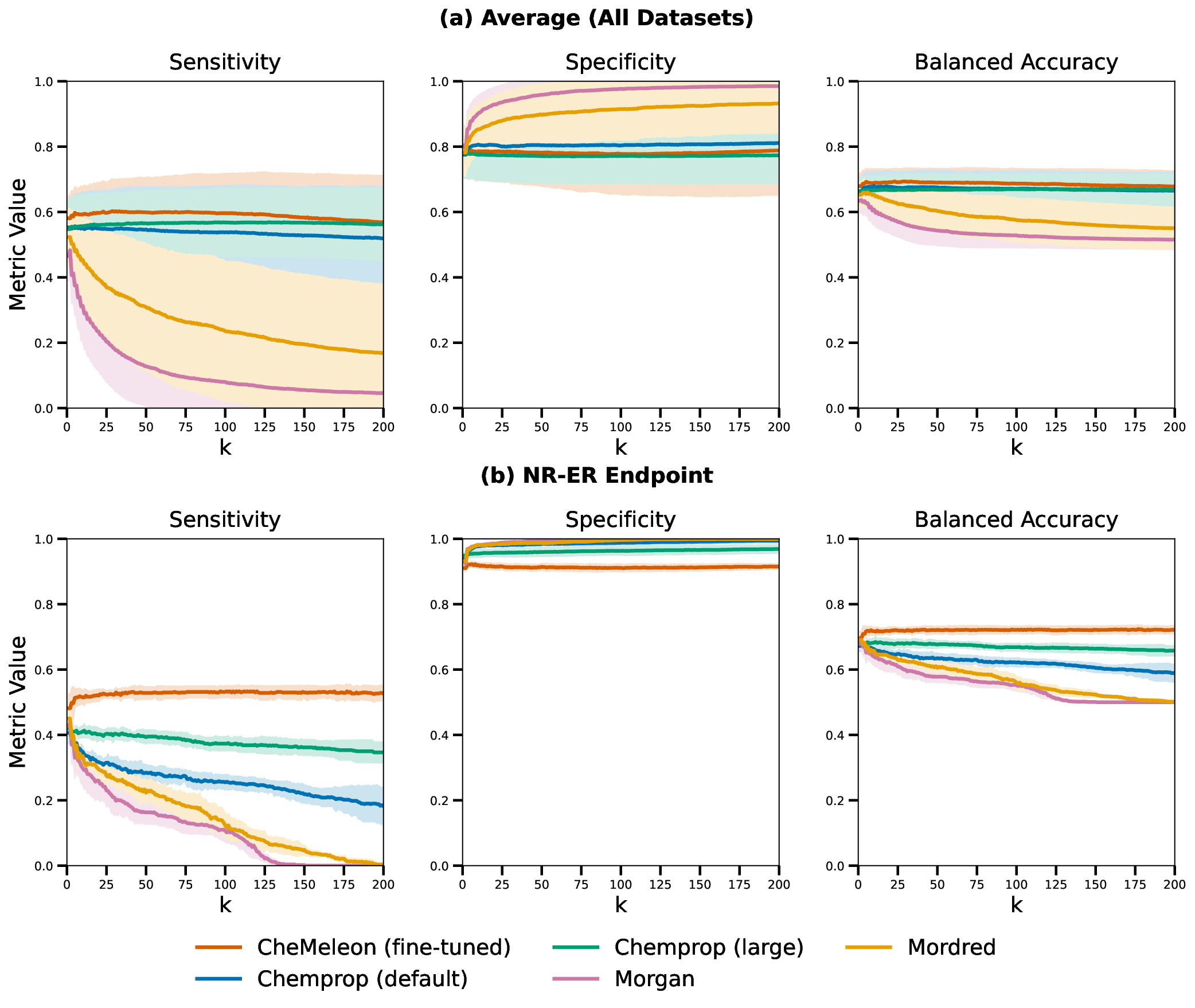}
    \caption{
    kNN probing of fixed and learned molecule representations.
    Upper row shows average results over 20 ToxCast endpoints.
    Lower row shows the results on the NR-ER endpoint, which was also used in the comparison by \citeauthor{ball2020key} \cite{ball2020key}.
    The results were obtained with a 5-fold cross validation using a random split, representing the read-across scenario.
    }
    \label{fig:neural_fp_random_split}
\end{figure}

The embeddings were learned by training or fine-tuning the models (\chemprop{} and \chemeleon{}) on the ToxCast training datasets and compared with fixed descriptor- and fingerprint-based representations.
Performance was quantified using balanced accuracy, sensitivity, and specificity, with results obtained from five-fold cross-validation using random and agglomerative cluster splits.
Figure~\ref{fig:neural_fp_random_split} summarizes the kNN probing results on a random split averaged across all endpoints, as well as results for Nuclear Receptor-Estrogen Receptor (NR-ER), an example endpoint measuring estrogen receptor activation, which is widely used in regulatory toxicology.
Similar trends are observed under agglomerative clustering-based splits (see Supporting Information Section \ref{molecular_representation_analysis}), also of interest in this domain.

\section{Discussion}\label{discussion} 

Our results show that pre-training on classical molecular descriptors enables the model to learn representations that not only reflect underlying chemical structure, but also translate into improved performance on practical property prediction tasks.
\chemeleon{}'s RMSE of 0.14 achieved during pre-training was mentioned in Section~\ref{results_pretraining} for completeness rather than its importance.
The model is never actually deployed to calculate these descriptors, especially given that they can be calculated exactly using \mordred{}.
While it is important that the pre-training performance improve continuously throughout training and thereby avoid overfitting, the importance of the actual performance metric is a question left for future work.

As shown in Figure \ref{fig:polaris_hsd_selected}, in many cases there are multiple winning models and in some cases, \textit{all} models achieve statistically indistinguishable performance, generally including \chemeleon{}.
\chemeleon{} narrowly outperforms the next best model \minimol{} on this subset of benchmarks, achieving a Win Rate of 75\%.
This demonstrates the effectiveness of molecular descriptors for pre-training.
Without the need for combining noisy experimental assays or running expensive QM simulations, \chemeleon{} is able to achieve state-of-the-art performance.

Also noteworthy is the relative strength of the Random Forest baseline model when compared against deep models when focusing on relevant benchmarks.
Many-parameter, well-known foundation models such as \molformer{} and \molclr{} fare surprisingly poorly.
\molformer{} reaches roughly equivalent performance to scratch-trained methods like \fastprop{} and \chemprop{}, while \molclr{} performs significantly worse.

Most promising here is the performance improvement that \chemeleon{} provides to \chemprop{}.
On its own, \chemprop{} falls below all but one of the tested methods, including the baseline Random Forest.
Given the ease of pre-training using the \chemeleon{} approach, it is reasonable to believe that future studies could use it to improve other model architectures in a similar manner.

When shifting focus to the MoleculeACE benchmarks, \chemeleon{} is statistically the best on 29 of the 30 benchmarks as shown in Table \ref{tab:model_performance_moleculeace}.
When narrowing in on the cliff molecules \chemeleon{} achieves a perfect win rate.
This is a dramatic improvement over the next-best \rf{} model, with only 15 wins and a 67\% win rate, respectively.
Like in the Polaris benchmark set, \chemeleon{} has again provided an enormous improvement over \chemprop{}.
Without the use of \chemeleon{} pre-training, the \chemprop{} model fails to achieve the highest performance in any of the test benchmarks.

While \chemeleon{}, \rf{}, and \minimol{} are the best-performing models overall, in most cases they fail to simultaneously achieve top overall performance and statistically similar performance on cliff and noncliff compounds.
These cases are indicated by the filled blue markers in Figure \ref{fig:mace_all}.
Across the tested benchmarks, \chemeleon{} does so only four times, while the next best model (\minimol{}) does so only twice.

The apparent difficulty of this benchmark for all tested models suggests that foundation modeling, while able to improve the absolute performance, does not reliably discern the presence of activity cliffs.
As shown in Supporting Information \ref{foundation_fingerprint}, \chemeleon{} learns a representation that maps structurally similar molecules into the same feature space, hampering efforts to distinguish them during fine-tuning on this task in particular.
Given the comparatively poor performance of the other tested foundation models \molformer{} and \minimol{}, this challenge seems to be endemic to the field.

Finally, in the kNN probing study \chemeleon{} achieves the highest average balanced accuracy among all evaluated representations across endpoints and exhibits a pronounced increase in sensitivity relative to both \chemprop{} and fixed representations.  
Because kNN predictions are determined solely by local neighborhood structure, this improvement indicates that CheMeleon organizes chemically and biologically similar compounds more effectively in the embedding space.  
The gain in sensitivity is achieved at only a modest reduction in specificity, yielding a favorable trade-off for toxicity prediction tasks.

We also sought to validate that performance improvements were not simply due to the increased number of parameters.
To that end, the ablated D‑MPNN \chemprop{} (large) is identical in size to \chemeleon{} but is trained from random initialization.
It performs substantially worse, particularly for small k.
This demonstrates that the improved embedding quality arises specifically from descriptor-based pre-training rather than architectural bias.
Similar trends are observed in this test when using an agglomerative clustering-based split (see Supporting Information Figure \ref{fig:neural_fp_agglomerative_split}) further affirming the applicability of \chemeleon{} to similarity driven applications in industrial and regulatory settings.

In conclusion, we have combined classical descriptor-based representations with state-of-the-art deep learning approaches to develop \chemeleon{}, a foundation model which provides statistically significant improvements to the underlying architecture relative to the best baselines in the literature.
\chemeleon{} is a powerful demonstration that its simple pre-training approach can improve existing modeling approaches and push the boundaries of chemical property prediction.
\chemeleon{} is open source, permissively licensed, and readily accessible via the \chemprop{} machine learning package \cite{chemprop_v2}.

\section{Methods}
\label{methods}

\subsection{\chemeleon{} Pre-training}
\label{chemeleon_pretraining}

Figure \ref{fig:workflow} visualizes the workflow for the present study.
The pre-training dataset was developed by randomly selecting 1 million molecules from PubChem \cite{pubchem} and calculating their molecular descriptors using \mordred{} \cite{mordred}.
Each descriptor represents a deterministic algorithm operating on the molecular graph.
These can be as simple as counting the presence of common functional groups and as complex as performing repeated shortest-path graph navigation, as in the Wiener index \cite{wiener_index}.
There are also parameterized descriptors based on externally estimated sub-properties, such as atomic volume, as used in the molecular McGowan Volume descriptor \cite{mcgowan_volume}.
The resulting dataset was stored in the Zarr format \cite{zarr} to enable efficient, high-throughput loading during training and has been made publicly available (see Section \ref{data_avail}).

For the model architecture, we employed a large D-MPNN implemented within the \chemprop{} framework \cite{chemprop_theory}.
The D-MPNN learns a molecular representation that is fed into a FNN to predict the computed descriptors.
To prevent overfitting and encourage robust feature learning, we applied a dynamic masking strategy where 85\% of the descriptor targets are randomly masked in the loss function.
The entire network was trained end-to-end to minimize the mean-squared error on the unmasked descriptors.

For downstream tasks, we initialize the model with the pre-trained D-MPNN weights while discarding the pre-trained FNN.
A fresh, task-specific FNN with random initial weights is then attached, and the entire model --- including the pre-trained encoder --- is fine-tuned end-to-end using stochastic gradient descent.
See Supporting Information Section \ref{training_details} for complete details on the model hyperparameters and training procedure.

\subsection{Benchmark Datasets}
\label{benchmarks}

The Polaris benchmarking suite is a collection of expert-curated datasets covering a broad range of molecular property prediction targets \cite{polaris}.
Users have also made available the datasets from the Therapeutic Data Commons (TDC) \cite{tdcommons}.
For this study, we selected a set of 28 unique benchmarks covering relevant targets such as solubility, physiology, and biophysics from both Polaris and TDC.
A subset of particularly relevant benchmarks whose performance across models is representative of the entire set has been shown in Figure~\ref{fig:polaris_hsd_selected}.
Each benchmark uses a different metric, automatically calculated by Polaris, all of which are either naturally scaled or rescaled (as done here) to range from zero to one, with one being the best.

The remaining 29 benchmarks are a set curated in the MoleculeACE study \cite{moleculeace}, each of which is an assay measuring the activity of small molecules against a biologically relevant target.
MoleculeACE evaluates model performance on activity cliffs.
Molecules are grouped into chemical series with small structural changes. 
Most changes lead to smooth activity variations (noncliff molecules), but occasionally a small modification causes a sharp decrease (i.e., "falls off a cliff") in activity; these cases are activity cliffs, and the corresponding compounds are labeled as cliff molecules.

\citeauthor{moleculeace} \cite{moleculeace} partition these chemical series in whole into either training or testing sets.
This allows calculation of the RMSE of test set activity predictions for the cliff and noncliff molecules.
These RMSEs are directly used for absolute performance comparisons.
The difference in RMSE between the cliff and noncliff molecules' predictions is checked via a one-sided, one-sample t-test ($\alpha=0.05$) to see if the model performs the same in both subgroups.
This is then summarized across all assays as the ``consistency rate'', a percentage reflecting the frequency with which the model maintains the same performance across activity cliffs.

Across all of these benchmarks we only perform single-task regression or classification.
Although multitask has been shown to improve performance in chemical datasets \cite{tetko_multitask}, we seek to demonstrate performance in individual tasks to better facilitate comparisons; all of the present models \textit{can} be used in a multitask fitting approach.
The present procedure enables us to quantify the performance of the models subject to the most significant source of uncontrollable, stochastic error.
Towards that same end, we do not perform hyperparameter optimization for any model on any benchmark dataset.
Such a procedure is known to lead to overfitting \cite{tetko_overfit}, particularly on the small datasets used in this study.

Although not central to the present work, we have provided performance results on the MoleculeNet benchmark collection \cite{moleculenet} to facilitate comparisons with historical models (see Supporting Information Section \ref{moleculenet}).
As discussed at length by \citeauthor{moleculenet_bad} \cite{moleculenet_bad}, many of the benchmarks included in this collection cover irrelevant or poorly defined tasks and contain data curation errors.
The physical chemistry, biophysics, and physiology categories from MoleculeNet are still represented in our selected benchmarks, but with the introduction of Polaris they are now more curated, standardized, and easier to access.

Finally, we sought a non-modeling-based evaluation method to provide foundational insights into the learned representation of \chemeleon{}.
To that end, we reproduce the ``Single Assay'' benchmark from \citeauthor{similarity_benchmark} \cite{similarity_benchmark} in Supporting Information Section \ref{foundation_fingerprint}, which enables quantitative comparisons of arbitrary molecular fingerprints' capacity to reproduce human-defined sorting order across thousands of series of small molecules.

\subsection{Reference Models}
\label{reference_models}

We compare \chemeleon{} against a diverse set of methods, ranging from classical baselines to state-of-the-art foundation models.
\begin{itemize}
    \item Random Forest \cite{sklearn, skmol}: A strong industrial baseline method. The present configuration (\rf{}) is trained on Morgan count fingerprints (dimension 2048, radius 2) \cite{Morgan1965, count_morgan} augmented with physicochemical descriptors from the RDKit \cite{rdkit}. Despite the prevalence of deep learning, such classical methods remain highly competitive and often outperform complex architectures in practical, low-data scenarios \cite{deeplose}, representing a realistic industrial standard.
    \item \fastprop{} \cite{fastprop}: A shallow FNN trained directly on \mordred{} descriptors \cite{mordred}. This serves to evaluate the predictive power of the descriptor set when used directly in a standard supervised learning framework.
    \item \chemprop{} \cite{chemprop_theory}: A standard D-MPNN initialized with random weights. This model serves as a direct baseline to evaluate the impact of our pre-training methodology on the same core architecture.
    \item \minimol{} \cite{minimol}: A parameter-efficient foundation model based on a GNN architecture similar to \chemprop{}. It is pre-trained in a multitask setting on the Graphium LargeMix dataset \cite{largemix}, where biological and quantum labels are sparsely annotated across molecules. It serves as a strong supervised learning baseline with a comparable parameter count to our method.
    \item \molformer{} \cite{molformer}: A large-scale chemical language model based on the Transformer architecture. Pre-trained on 1.1 billion SMILES strings via masked language modeling, it represents the state-of-the-art in sequence-based representations.
    \item \molclr{} \cite{molclr}: A foundation model utilizing a GNN backbone to learn generalized representations via contrastive learning on 2D molecular graphs without any labels.
\end{itemize}

These models represent the state-of-the-art in learning from graphs, text, and descriptors, with Random Forest included to provide a strong classical baseline.

\subsection{Model Comparisons}
\label{statistical_comparisons}

For all benchmarks, we use a single fixed test set as defined by the original dataset creators.
This approach ensures compatibility with  previous studies and facilitates rigorous statistical comparisons between our tested models.
To generate the necessary replicate measurements for statistical analysis, we adopt the following procedure for each benchmark:
\begin{enumerate}
    \item Select a random seed for the current repetition.
    \item Based on the model architecture, use this random seed to:
    \begin{enumerate}
        \item (for deep models) define the validation split for early stopping to prevent overfitting.
        \item (for \rf{}) initialize the feature bagging.
    \end{enumerate}
    \item Train the model on the training set, using the validation set for early stopping where applicable.
    \item Evaluate the trained model on held-out test set.
\end{enumerate}

We compare the performance of all models simultaneously using the Tukey Honestly Significant Difference (HSD) test \cite{statsmodels, polaris} ($\alpha=0.05$).
This procedure controls the family-wise error rate to account for multiple comparisons.
For a given benchmark, the top-performing model and any models statistically indistinguishable from it are designated as winners.
This is aggregated across all benchmarks to arrive at a win count and win rate.

\subsection{k-Nearest Neighbors Representation Probing}
\label{rep_analysis}

An effective molecular representation should encode chemically meaningful structure such that distances in the embedding space reflect molecular similarity and, ideally, similarity in molecular properties.
To evaluate the suitability of different molecular representations, we employ k-Nearest Neighbor (kNN) probing~\cite{praski2025benchmarking, wu2018unsupervised, caron2021emerging}.
This analysis compares embeddings learned by neural models such as \chemeleon{} and \chemprop{} with traditional fixed molecular representations, including count-based Morgan fingerprints and physicochemical descriptors as used in \rf{}.

We applied kNN probing to 20 challenging endpoints (balanced accuracies between $\sim$0.6 and 0.8) from ToxCast Database, reflecting real-world scenarios~\cite{feldmann2025analysis}. 
The ToxCast Database is a public resource containing \textit{in vitro} screening assay data produced by the U.S. Environmental Protection Agency’s Toxicity Forecaster program \cite{richard2016toxcast}.
In this work, we used the ToxCast datasets made available via  MoleculeNet \cite{moleculenet}.

In addition, we followed the read-across evaluation protocol of \citeauthor{ball2020key} \cite{ball2020key} for all the endpoints including NR-ER.
These assays measure a compound’s potential to interact with the estrogen receptor, a key mechanism in endocrine disruption and regulatory toxicology.
Read-across is a similarity-based inference strategy in which the activity of a query compound is inferred from structurally or chemically similar compounds, making it particularly sensitive to the quality of the underlying molecular representation.

For \chemeleon{} and \chemprop{}, molecular embeddings were obtained by first fine-tuning on the training set and then generating embeddings for both train and test sets for kNN analysis.
Compound preprocessing, descriptor calculation, and neural fingerprint extraction followed the procedures described in \citeauthor{feldmann2025analysis}~\cite{feldmann2025analysis}.
All experiments were conducted using five-fold cross-validation under both random and agglomerative clustering-based splits.

\backmatter

\section*{Data Availability}
\label{data_avail}
All datasets used in this work are available from their respective authors via Polaris or GitHub.
Molecule descriptors calculated from the publicly available PubChem structures have been made available on Zenodo with accession code 10.5281/zenodo.15733574 \cite{chemeleon_training_data}.
The \chemeleon{} model weights have been made available on Zenodo with accession code 10.5281/zenodo.15426600 \cite{chemeleon_weights}.
Code for automatic downloading and usage of weights is available with the source code and has been integrated into the \chemprop{} software package as of version 2.2.0.

\section*{Code Availability}
\label{code_avail}
The source code associated with this study, including pre-training, fine-tuning, numerical results, and visualization of the same, is available at \href{https://github.com/JacksonBurns/CheMeleon}{GitHub.com/JacksonBurns/CheMeleon}.

\newpage
\bibliography{main}

\section*{Acknowledgments}

JWB acknowledges that this material is based upon work supported by the U.S. Department of Energy, Office of Science, Office of Advanced Scientific Computing Research, Department of Energy Computational Science Graduate Fellowship under Award Number DE-SC0023112. 
ASZ acknowledges the fellowship support from MathWorks.
CRAA and WHG gratefully acknowledge financial support from the Machine Learning for Pharmaceutical Discovery and Synthesis consortium.
Finally, the authors thank Shih-Cheng Li, Prof. Connor Coley, and Prof. Markus Kraft for their valuable comments and suggestions.

\section*{Author Contributions Statement}

JWB: conceptualization (equal); methodology (equal); formal analysis (equal); software (lead); writing - original draft (lead); review and editing (equal). ASZ: conceptualization (equal); methodology (equal); formal analysis (equal); software (supporting); writing - original draft (supporting); review and editing (equal). CRAA: methodology (supporting); software (supporting); formal analysis (supporting); writing - original draft (supporting); review and editing (equal). JS: formal analysis (supporting); software (supporting); writing - original draft (supporting); review and editing (supporting). CF: formal analysis (supporting); software (supporting); writing - original draft (supporting); review and editing (supporting). MM: formal analysis (supporting); software (supporting); writing - original draft (supporting); review and editing (supporting). WHG: supervision (lead); formal analysis (supporting); review and editing (supporting).

\section*{Competing Interests Statement}
The authors declare no competing interests.

\end{document}